\PassOptionsToPackage{table}{xcolor}
\PassOptionsToPackage{authoryear,round}{natbib}
\documentclass[10pt,a4paper,logo]{googledeepmind}
\usepackage{times}
\usepackage{natbib}

\usepackage{amsmath,amsfonts,bm}

\def\eqref#1{equation~\ref{#1}}

\def\1{\bm{1}}

\DeclareMathAlphabet{\mathsfit}{\encodingdefault}{\sfdefault}{m}{sl}
\SetMathAlphabet{\mathsfit}{bold}{\encodingdefault}{\sfdefault}{bx}{n}

\usepackage{url}
\usepackage{graphicx}
\usepackage{wrapfig}
\usepackage{booktabs}
\usepackage[table]{xcolor}
\usepackage{tabularx}
\definecolor{settinggray}{HTML}{F2F3F5}
\definecolor{bestcell}{HTML}{9CCDB5}
\definecolor{secondcell}{HTML}{E3F2E9}
\newcommand{\metricbest}[1]{\cellcolor{bestcell}{\bfseries\boldmath #1}}
\newcommand{\metricsecond}[1]{\cellcolor{secondcell}#1}
\newcommand{\metricvalue}[1]{{\fontsize{8.5}{10}\selectfont #1}}
\newcommand{\metricpm}[2]{\mbox{\metricvalue{#1}{\fontsize{5}{6}\selectfont$\pm$#2}}}
\usepackage{amsmath,amssymb}
\usepackage{algorithm}
\usepackage{algpseudocode}
\usepackage{hyperref}

\title{Teacher--Student Gaps Are Not Enough: Outcome-Guided On-Policy Distillation for Multi-Turn Autonomous Agents}
\author[1,2]{Tong~Zhang\textsuperscript{*}}
\author[2]{Zhou~Liu}
\author[1]{Yihao~Liu\textsuperscript{*}}
\author[1]{Jiahua~Bao\textsuperscript{*}}
\author[3]{Xuchen~Li}
\author[4]{Honglin~Lin}
\author[1]{Tao~Cheng\textsuperscript{*}}
\author[1]{Zhihan~Yu\textsuperscript{*}}
\author[1]{Kai~Tang\textsuperscript{\dag}}
\author[1]{Xiaoxi~Jiang}
\author[1]{Guanjun~Jiang}
\affil[1]{Qwen Large Model Application Team, Alibaba}
\affil[2]{Peking University}
\affil[3]{University of Chinese Academy of Sciences}
\affil[4]{Shanghai~Jiao~Tong~University}
\correspondingauthor{haoting.tk@alibaba-inc.com}
\hypersetup{pdftitle={Teacher--Student Gaps Are Not Enough: Outcome-Guided On-Policy Distillation for Multi-Turn Autonomous Agents},pdfauthor={Tong Zhang, Zhou Liu, Yihao Liu, Jiahua Bao, Xuchen Li, Honglin Lin, Tao Cheng, Zhihan Yu, Kai Tang, Xiaoxi Jiang, Guanjun Jiang}}

\newcommand{\sg}{\operatorname{sg}}
\newcommand{\ind}{\mathbb{I}}
\newcommand{\pos}[1]{\left[#1\right]_{+}}
\AddToHook{cmd/normalsize/after}{%
  \setlength{\abovedisplayskip}{6pt plus 1pt minus 1pt}%
  \setlength{\belowdisplayskip}{6pt plus 1pt minus 1pt}%
  \setlength{\abovedisplayshortskip}{2pt plus 1pt}%
  \setlength{\belowdisplayshortskip}{4pt plus 1pt minus 1pt}%
}

\begin{abstract}
On-policy distillation (OPD) trains a student on its own trajectories with dense teacher supervision. Recent work on OPD for multi-turn autonomous agents often treats large teacher--student token-level distributional gaps as promising intervention points, linking larger gaps to a greater need for correction. Yet, our empirical analysis reveals a \textbf{supervision--benefit mismatch}: large gaps can be benign, while small gaps can be outcome-critical. Teacher--student gaps capture differences at the current turn, whereas the benefit of teacher guidance depends on how the current student interacts with the environment afterward. The student may still succeed despite choosing an action that differs from the teacher's, while a teacher-preferred action may lead to a state from which the student cannot complete the task. Local gaps alone are therefore not enough to determine whether teacher guidance benefits the current student. Effective supervision should instead emphasize guidance that the current student can translate into better final task outcomes. Accordingly, we propose \textbf{Outcome-Guided On-Policy Distillation (OG-OPD)}, which applies trajectory-relative weighting to teacher supervision and calibrates these weights using final task outcomes from paired student continuations. This calibration selectively strengthens supervision on the student's original trajectories at turns where teacher guidance benefits the current student. Across ALFWorld, ScienceWorld, and WebShop, OG-OPD consistently outperforms baselines under diverse settings. It improves task success rates by 3.6--17.7 percentage points over vanilla OPD and by up to 7.0 percentage points over the strongest baseline.
\end{abstract}

\begin{document}
\maketitle

\raggedbottom
\setlength{\textfloatsep}{14pt plus 1pt minus 1pt}
\begin{figure}[!b]
\centering
\includegraphics[width=\linewidth]{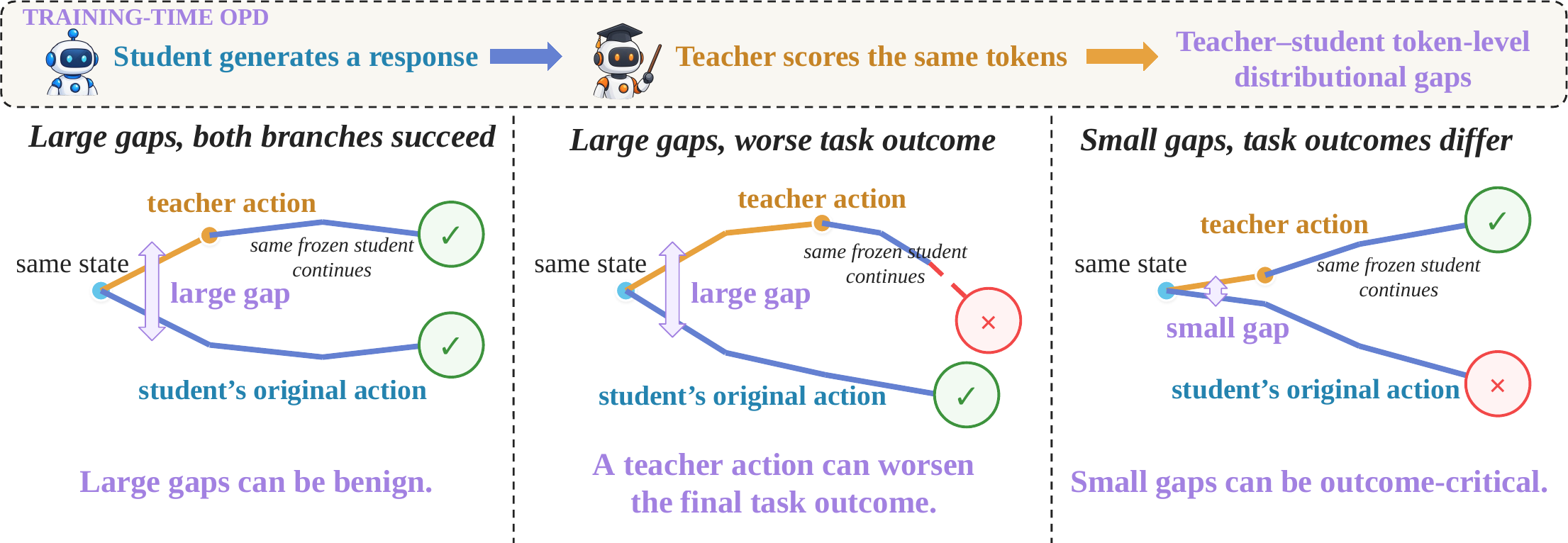}%
\caption{\textbf{Illustration of the supervision--benefit mismatch.} From the same state and interaction history, one branch takes the student's original action and the other a teacher-proposed action. Both branches then continue interacting under the same frozen student policy. Moving toward the teacher's preferred action does not necessarily improve the current student's final task outcome.}
\label{fig:motivation}
\end{figure}\section{Introduction}

On-policy distillation (OPD) trains students on their own trajectories \citep{agarwal2024onpolicy,lin2020autoregressive} with dense teacher supervision \citep{lu2025onpolicy,fu2026rethinking}. Learning at student-visited states helps reduce the distribution mismatch between training and inference \citep{bengio2015scheduled,ranzato2016sequence}. At each turn, the teacher provides token-level probability feedback on the student's response \citep{jin2026entropy,ko2024distillm,jia2026asymmetric}, encouraging the student to narrow teacher--student token-level distributional gaps \citep{gu2024minillm,jang2026stable,shao2026token}. However, local teacher--student gaps alone are not enough to determine whether teacher guidance benefits the current student. This limitation is especially important for multi-turn autonomous agents, where teacher supervision spans successive environment interactions \citep{wang2026tcod,liao2026prefix}. Actions at one turn shape subsequent observations \citep{yao2023react} and the states in which later decisions are made \citep{ross2011dagger}. Errors can compound across successive turns \citep{ross2010efficient,li2026guided}. Effective supervision must therefore consider both where to intervene and what the current student achieves afterward.

Against this background, existing agentic OPD methods organize teacher supervision across turns through rollout construction and teacher intervention. Some progressively extend student rollouts or shorten prefixes of successful teacher trajectories \citep{wang2026tcod}, giving the student control over more turns as training progresses. Others interleave teacher and student turns within a trajectory while gradually reducing the probability of teacher intervention \citep{li2026guided}. For more targeted guidance, recent work uses teacher--student token-level distributional gaps to identify promising intervention points and steer subsequent student behavior toward the teacher \citep{chen2026futurebridge}. Yet, does teacher guidance at these turns actually improve the current student's final task outcome?

To examine this relationship, we conduct an empirical analysis on ScienceWorld \citep{wang2022scienceworld} and WebShop \citep{yao2022webshop}. Our findings reveal a \textbf{supervision--benefit mismatch}: large gaps can be benign, while small gaps can be outcome-critical. A teacher-proposed action can change subsequent states and decisions, but its benefit depends on the current student's ability to complete the task from the resulting state. Assessing this benefit requires comparing the current student's final task outcomes after its original action and after the teacher action. Local gaps remain useful for locating potential intervention points, while this comparison provides evidence for deciding which turns warrant stronger teacher supervision. Figure~\ref{fig:motivation} shows a schematic, with details in Section~\ref{sec:diagnosis}.

Accordingly, we propose \textbf{Outcome-Guided On-Policy Distillation (OG-OPD)}. First, OG-OPD uses trajectory-relative weighting to allocate teacher supervision and identifies candidate turns based on the current gap and an abrupt increase from the preceding turn. It then calibrates these weights using final task outcomes from paired student continuations. When the teacher action improves the current student's final task outcome, OG-OPD selectively strengthens supervision at the corresponding turn on the student's own trajectory. Otherwise, it retains the pre-calibration weight. Teacher responses and their continuations are used only for calibration, not as additional distillation targets.

We evaluate OG-OPD on ALFWorld \citep{shridhar2021alfworld}, ScienceWorld, and WebShop under different teacher--student configurations. OG-OPD achieves the highest mean task success rate among the compared methods in every evaluated setting. On WebShop, for example, the Qwen3-1.7B student achieves a 45.7\% success rate, compared with 28.0\% for vanilla OPD and 38.7\% for the strongest competing baseline. These gains extend to higher task scores on both ScienceWorld and WebShop. OG-OPD also requires fewer interaction turns than vanilla OPD in most comparisons.

Our main contributions are as follows. (1) \textbf{Supervision--benefit mismatch.} Moving toward the teacher does not necessarily benefit the current student. Across the two benchmarks, teacher harm averages 10.8\% of large-gap decisions, exceeding teacher rescue at 8.8\%. (2) \textbf{Outcome-Guided On-Policy Distillation.} OG-OPD builds on trajectory-relative weighting and uses task outcomes to decide where to strengthen teacher supervision. (3) \textbf{Broad validation.} Experiments on three benchmarks with different teacher--student settings show consistent gains in task success over baselines.

\section{Preliminaries}
\label{sec:preliminaries}
\subsection{Multi-turn Interaction}
\begingroup
\setlength{\abovedisplayskip}{3.5pt plus 1pt minus 1pt}
\setlength{\belowdisplayskip}{3.5pt plus 1pt minus 1pt}
\setlength{\abovedisplayshortskip}{1pt plus 1pt}
\setlength{\belowdisplayshortskip}{3pt plus 1pt minus 1pt}
At turn $k$, the frozen rollout student $p_\circ$ generates response $z_k$ from history $c_k$. The action parser $\mathcal A$ maps $z_k$ to action $u_k$, which changes environment state $s_k$ according to transition law $\mathsf T_{\mathrm{env}}$:
\begin{equation}
\begin{aligned}
z_k&\sim p_\circ(\cdot\mid c_k),\qquad u_k=\mathcal A(z_k),\\
(s_{k+1},o_{k+1})&\sim \mathsf T_{\mathrm{env}}(\cdot\mid s_k,u_k),\qquad c_{k+1}=c_k\oplus(z_k,o_{k+1}),
\end{aligned}
\label{eq:interaction}
\end{equation}
Here $s_k$ need not be fully observable and $\oplus$ appends the response and observation. Autoregressive generation gives $p_\circ(z_k\mid c_k)=\prod_j p_\circ(z_{k,j}\mid v_{k,j})$, with $v_{k,j}=(c_k,z_{k,<j})$. Trajectory $\zeta$ ends at termination or the interaction limit. Its final task outcome $\mathsf S(\zeta)\in\{0,1\}$ indicates benchmark success. Turn sums cover $0\le k<K_\zeta$, where $K_\zeta$ counts the interaction turns in trajectory $\zeta$.
\par\endgroup

\raggedbottom
\setlength{\textfloatsep}{11pt plus 1pt minus 1pt}
\setlength{\floatsep}{5pt plus 1pt minus 1pt}
\setlength{\intextsep}{10pt plus 1pt minus 1pt}
\subsection{On-policy Distillation}
\label{sec:opd}
\begingroup
\setlength{\abovedisplayskip}{3.5pt plus 1pt minus 1pt}
\setlength{\belowdisplayskip}{3.5pt plus 1pt minus 1pt}
\setlength{\abovedisplayshortskip}{1pt plus 1pt}
\setlength{\belowdisplayshortskip}{3pt plus 1pt minus 1pt}
OPD provides dense teacher supervision on the student's own trajectories \citep{agarwal2024onpolicy}. For fixed teacher $q_\phi$ and trainable student $p_\theta$, reverse KL distillation \citep{gu2024minillm} minimizes
\begin{equation}
\begin{aligned}
\mathcal J_{\mathrm{rKL}}(\theta)
&=\mathbb E_{v\sim d_{p_\circ}}\!\left[D_{\mathrm{KL}}\!\left(p_\theta(\cdot\mid v)\,\|\,q_\phi(\cdot\mid v)\right)\right],\\
D_{\mathrm{KL}}\!\left(p_\theta(\cdot\mid v)\,\|\,q_\phi(\cdot\mid v)\right)
&=\sum_{x\in\mathcal X}p_\theta(x\mid v)\log\frac{p_\theta(x\mid v)}{q_\phi(x\mid v)},
\end{aligned}
\label{eq:reverse-kl}
\end{equation}
where $\mathcal X$ is the vocabulary. The token context distribution $d_{p_\circ}$ is sampled under the frozen rollout student $p_\circ$ and held fixed while the trainable student $p_\theta$ is optimized within each update.

For each sampled token $z_{k,j}$, the teacher supervision and importance ratio are
\begin{equation}
\psi_{k,j}=\log\frac{q_\phi(z_{k,j}\mid v_{k,j})}{p_\circ(z_{k,j}\mid v_{k,j})},
\qquad
\rho_{k,j}(\theta)=\frac{p_\theta(z_{k,j}\mid v_{k,j})}{p_\circ(z_{k,j}\mid v_{k,j})}.
\label{eq:signal}
\end{equation}
At fixed $v$, the expected signed log probability ratio under $p_\circ$ equals the negative reverse KL:
\begin{equation}
\mathbb E_{x\sim p_\circ(\cdot\mid v)}\!\left[\log\frac{q_\phi(x\mid v)}{p_\circ(x\mid v)}\right]
=-D_{\mathrm{KL}}\!\left(p_\circ(\cdot\mid v)\,\|\,q_\phi(\cdot\mid v)\right).
\label{eq:sampled-kl}
\end{equation}
Positive $\psi_{k,j}$ favors a higher sampled token probability, while negative values favor a lower one.

Let $\mathcal M_k(\zeta)$ index generated tokens with valid teacher scores, excluding prompts, observations, and padding, and let $Z=\sum_{\zeta\in\mathcal D}\sum_k|\mathcal M_k(\zeta)|$ count them in batch $\mathcal D$. The clipped policy surrogate \citep{schulman2017ppo}, defined in Appendix~\ref{app:optimization}, gives the turn-level and batch losses to minimize:
\begin{equation}
\begin{aligned}
\mathcal Q_k(\theta;\zeta)
&=\sum_{j\in\mathcal M_k(\zeta)}
\ell_{\mathrm{pol}}\!\left(\rho_{k,j}(\theta),\sg[\psi_{k,j}]\right),\\
\mathcal L_{\mathrm{OPD}}(\theta)
&=\frac{1}{Z}\sum_{\zeta\in\mathcal D}\sum_k\mathcal Q_k(\theta;\zeta).
\end{aligned}
\label{eq:opd-objective}
\end{equation}
Here $\sg[\cdot]$ holds $\psi_{k,j}$ fixed during differentiation. OG-OPD changes only the turn-level weights.

We quantify the magnitude of teacher--student token-level distributional gaps at each turn using
\begin{equation}
\chi_k=\frac{1}{|\mathcal M_k(\zeta)|}\sum_{j\in\mathcal M_k(\zeta)}|\psi_{k,j}|,\qquad k\in\mathcal V(\zeta).
\label{eq:response-gap}
\end{equation}
Here $\mathcal V(\zeta)=\{k:|\mathcal M_k(\zeta)|>0\}$ retains original turn indices. The sampled gap magnitude $\chi_k$ is distinct from signed token-level update directions and full-vocabulary KL divergence.
\par\endgroup

\section{Outcome-Guided On-Policy Distillation}
\label{sec:method}
We first present an empirical analysis of the supervision--benefit mismatch. Motivated by these findings, we propose OG-OPD, building on trajectory-relative weighting and candidate selection. Its core component, outcome-based calibration, uses final task outcomes from paired student continuations to selectively strengthen teacher supervision on the student's own trajectories.

\subsection{Empirical Analysis}
\label{sec:diagnosis}
\begingroup
\setlength{\intextsep}{8pt}
We analyze paired student continuations on ScienceWorld and WebShop using a fixed Qwen3-32B teacher and Qwen3-1.7B student. At sampled student-visited states, one branch takes the student's original action and the other the teacher action, with both starting from the same state and history and continuing under the same frozen student policy. Within each benchmark, the highest and lowest of five quantile groups define the large-gap and small-gap groups. Each decision is classified from five matched trials. Appendix~\ref{app:diagnostics} describes state sampling and the full outcome classification rules.

\setlength{\columnsep}{10pt}
\begin{wrapfigure}{r}{0.50\linewidth}
\centering
\setlength{\abovecaptionskip}{0pt}
\setlength{\belowcaptionskip}{0pt}
\includegraphics[width=\linewidth]{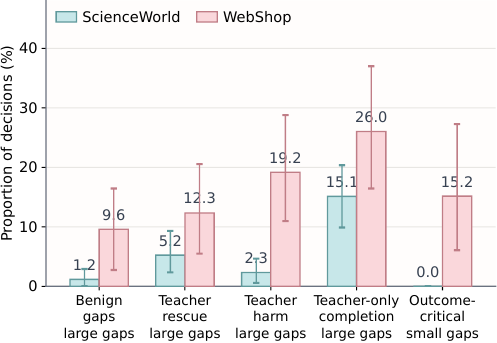}
\caption{\textbf{Empirical analysis of final task outcomes.} Bars show decision proportions by outcome category within each benchmark's gap groups. Error bars indicate 95\% decision-level bootstrap confidence intervals.}
\label{fig:outcomes}
\end{wrapfigure}
Figure~\ref{fig:outcomes} summarizes final task outcomes across gap groups. In the large-gap group, benign gaps account for an average of 5.4\% of decisions, where both the student's original action and the teacher action lead to success. Teacher rescue and teacher harm occur in 8.8\% and 10.8\% of decisions, respectively. These findings indicate that a large gap can be benign, while the teacher action can either improve or worsen the final task outcome. In the small-gap group, 7.6\% of decisions are outcome-critical, with the two branches leading to different final task outcomes. Together, these results reveal the supervision--benefit mismatch. Beyond these averages, the relative frequencies of teacher rescue and teacher harm vary across benchmarks. Within the large-gap group, teacher harm occurs more often than teacher rescue on WebShop, whereas the reverse holds on ScienceWorld. Appendix~\ref{app:selection} further examines this pattern under gap magnitude ranking.

Beyond gap magnitude, we examine the role of the continuation policy by comparing teacher and student continuations after the same teacher action. In an average of 20.6\% of large-gap decisions across the two benchmarks, the teacher frequently completes the task, while the current student rarely succeeds from the same resulting state. We refer to these cases as teacher-only completion. This suggests that the value of teacher guidance depends on the current student's ability to continue from the resulting state and complete the remaining task successfully. These findings motivate OG-OPD to use paired student continuations when deciding where to strengthen teacher supervision.
\par\endgroup

\subsection{Trajectory-Relative Weighting and Candidate Selection}
\label{sec:base-selection}
\begingroup
\setlength{\abovedisplayskip}{3.5pt plus 1pt minus 1pt}
\setlength{\belowdisplayskip}{3.5pt plus 1pt minus 1pt}
\setlength{\abovedisplayshortskip}{1pt plus 1pt}
\setlength{\belowdisplayshortskip}{3pt plus 1pt minus 1pt}
Figure~\ref{fig:method} illustrates trajectory-relative weighting and candidate selection. For the gap $\chi_k$ in Equation~\ref{eq:response-gap}, we define its logarithmic magnitude $\nu_k$ and change from the preceding turn $\upsilon_k$ as
\begin{equation}
\nu_k=\log(\varepsilon_0+\chi_k),
\qquad
\upsilon_k=\nu_k-\nu_{k-1}
=\log\frac{\varepsilon_0+\chi_k}{\varepsilon_0+\chi_{k-1}},
\label{eq:gap-signals}
\end{equation}
where $\varepsilon_0>0$ stabilizes the logarithm. The change $\upsilon_k$ requires adjacent valid turns $k,k-1\in\mathcal V(\zeta)$.

Since $\chi_k$ averages $|\psi_{k,j}|$, larger gaps imply larger supervision coefficients on average. Trajectory-relative weighting moderates this variation using the first valid turn $k_0=\min\mathcal V(\zeta)$ as a reference. Each valid turn in the student's own trajectory receives the pre-calibration weight
\begin{equation}
\beta_k=
\begin{cases}
1,&k=k_0,\\
\min\!\left\{\beta_{\max},\exp(\nu_{k_0}-\nu_k)\right\},&k>k_0,
\end{cases}
\qquad \beta_{\max}\ge1.
\label{eq:base}
\end{equation}
Gaps above the reference reduce the weight, while gaps below it increase the weight up to $\beta_{\max}$. Using Equation~\ref{eq:response-gap}, the average magnitude of the weighted token-level teacher supervision is
\begin{equation}
\frac{1}{|\mathcal M_k(\zeta)|}\sum_{j\in\mathcal M_k(\zeta)}|\beta_k\psi_{k,j}|
=\beta_k\chi_k
=\min\!\left\{\beta_{\max}\chi_k,\frac{(\varepsilon_0+\chi_{k_0})\chi_k}{\varepsilon_0+\chi_k}\right\}.
\label{eq:weighted-gap}
\end{equation}
With $\varepsilon_0$ negligible relative to the gaps and the cap inactive, $\beta_k\chi_k\approx\chi_{k_0}$. This sets a common supervision scale across valid turns in the trajectory, but does not determine whether teacher guidance benefits the current student in terms of its final task outcome under its continuation policy.

To limit additional environment execution, we evaluate paired student continuations only at selected turns. Candidates require a sufficiently large gap and, after the initial turn, a positive increase meeting its threshold. Final task outcomes determine whether to strengthen supervision. Let $\mathcal H(\zeta)$ contain turns with valid scores and replay information. We define the candidate set as
\begin{equation}
\mathcal K(\zeta)=\left\{k\in\mathcal H(\zeta)\;\middle|\;
\nu_k\ge \vartheta_\nu\ \land\
\bigl[k=0\ \lor\ (\upsilon_k>0\ \land\ \upsilon_k\ge \vartheta_\upsilon)\bigr]\right\}.
\label{eq:candidates}
\end{equation}
Thresholds are batch quantiles with equal total weight per contributing trajectory. Only positive valid changes enter $\vartheta_\upsilon$. The magnitude-only exception applies to $k=0\in\mathcal H(\zeta)$, not a later first valid turn $k_0>0$. Within a bounded proposal budget, we check candidates chronologically and select the first replay-valid turn $k^\star$ whose teacher response yields a legal action different from the student's original action. Only this turn undergoes paired student continuations for outcome-based calibration in Section~\ref{sec:calibration}. Appendix~\ref{app:implementation} gives further details of this execution procedure.

\begin{figure}[!t]
\centering
\includegraphics[width=\linewidth]{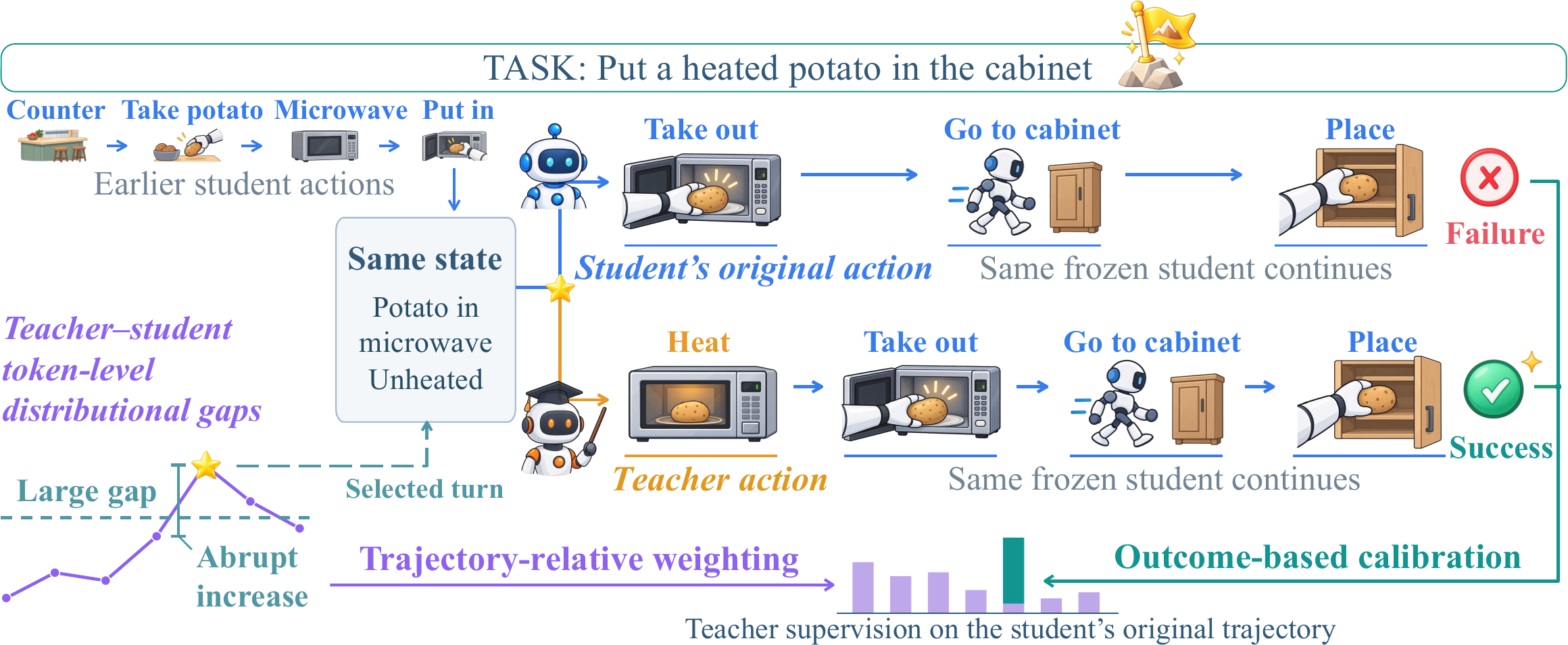}%
\caption{\textbf{Overview of Outcome-Guided On-Policy Distillation.} Trajectory-relative weighting assigns supervision weights, while gap magnitude and abrupt increases identify candidate turns. Outcome-based calibration compares paired student continuations and strengthens supervision on the student's original trajectory when teacher guidance improves the final task outcome.}
\label{fig:method}
\end{figure}
\par\endgroup

\subsection{Outcome-Based Calibration}
\label{sec:paired}
\label{sec:calibration}
\begingroup
\setlength{\abovedisplayskip}{3.5pt plus 1pt minus 1pt}
\setlength{\belowdisplayskip}{3.5pt plus 1pt minus 1pt}
\setlength{\abovedisplayshortskip}{1pt plus 1pt}
\setlength{\belowdisplayshortskip}{3pt plus 1pt minus 1pt}
Outcome-based calibration increases the selected candidate turn's pre-calibration weight only when the current student fails with its original action but succeeds with the teacher action.

Let $z_{k^\star}^{S}=z_{k^\star}$ denote the student's original response and $z_{k^\star}^{T}$ the selected teacher response sampled from $q_\phi(\cdot\mid c_{k^\star})$. We restore state $s_{k^\star}$ and history $c_{k^\star}$ from the student's own trajectory. Define $\mathcal U(s,c,z;p_\circ,\varsigma)$ to execute $\mathcal A(z)$, retain $z$ in history, and continue under the frozen current student $p_\circ$ with seed schedule $\varsigma$. The paired student continuations are then generated as follows
\begin{equation}
\begin{aligned}
\tilde\zeta^{S}&=\mathcal U(s_{k^\star},c_{k^\star},z_{k^\star}^{S};p_\circ,\varsigma),\\
\tilde\zeta^{T}&=\mathcal U(s_{k^\star},c_{k^\star},z_{k^\star}^{T};p_\circ,\varsigma).
\end{aligned}
\label{eq:pair}
\end{equation}
Superscript $T$ denotes the teacher response, not teacher continuation. Both use the same frozen current student, initial state, history, decoding settings, remaining interaction budget, and seed schedule. Responses may change subsequent states, so we test whether the current student can complete the task after taking the teacher action, rather than whether the teacher can. Training uses one pair at the selected turn, whereas the empirical analysis repeats matched trials per sampled decision.

Let $\mathcal D_{\mathrm{pair}}\subseteq\mathcal D$ contain trajectories with valid completed paired student continuations. Define
\begin{equation}
\Delta_\zeta=\mathsf S(\tilde\zeta^{T})-\mathsf S(\tilde\zeta^{S})\in\{-1,0,1\}.
\label{eq:outcome-difference}
\end{equation}
Here $\Delta_\zeta=1$ denotes failure under the original action and success under the teacher action. $\Delta_\zeta=-1$ denotes the reverse, and $\Delta_\zeta=0$ denotes no change in final task outcome. Only the positive case permits additional distillation weight at the selected candidate turn.
\begin{equation}
\mathsf g_\zeta=\mathsf S(\tilde\zeta^{T})\bigl(1-\mathsf S(\tilde\zeta^{S})\bigr)
=\ind[\Delta_\zeta=1]=\pos{\Delta_\zeta}.
\label{eq:gate}
\end{equation}
Success with the teacher action provides positive execution evidence only when the original action fails, excluding cases where the current student succeeds under both responses.

The calibrated weight $\omega_k$ adds the outcome-based increment $\gamma_\zeta$ to the pre-calibration weight at the selected turn. For weight floor $\beta_{\uparrow}>0$ and $[x]_+=\max\{x,0\}$, the resulting weights are
\begin{equation}
\gamma_\zeta=\mathsf g_\zeta\pos{\beta_{\uparrow}-\beta_{k^\star}},
\qquad
\omega_k=
\begin{cases}
\beta_k+\gamma_\zeta,&k=k^\star,\\
\beta_k,&k\ne k^\star.
\end{cases}
\label{eq:weight}
\end{equation}
When $\mathsf g_\zeta=1$, the calibrated weight is $\max\{\beta_{k^\star},\beta_{\uparrow}\}$. Paired outcomes determine whether to add supervision, while the floor and pre-calibration weight determine the increase. Positive execution evidence can thus strengthen supervision at a previously downweighted turn. Otherwise, or without valid paired student continuations, $\gamma_\zeta=0$ and pre-calibration weights remain unchanged. Since $\beta_{\max}$ is not reapplied, a higher floor can raise the calibrated weight above the pre-calibration cap.

For a fixed batch $\mathcal D$ of the student's own trajectories, calibrated weights scale the original response losses $\mathcal Q_k(\theta;\zeta)$ from Section~\ref{sec:opd}. Teacher responses and paired student continuations are used only for outcome-based calibration, not as additional distillation targets. Keeping the original token normalization $Z$ unchanged, we write the resulting weighted training objective as
\begin{equation}
\begin{aligned}
\mathcal L_{\mathrm{OG-OPD}}(\theta)
&=\frac{1}{Z}\sum_{\zeta\in\mathcal D}\sum_{k\in\mathcal V(\zeta)}\sg[\omega_k]\,\mathcal Q_k(\theta;\zeta)\\
&=\underbrace{\frac{1}{Z}\sum_{\zeta\in\mathcal D}\sum_{k\in\mathcal V(\zeta)}\sg[\beta_k]\,\mathcal Q_k(\theta;\zeta)}_{\mathcal L_{\mathrm{base}}(\theta)}
+\frac{1}{Z}\sum_{\zeta\in\mathcal D_{\mathrm{pair}}}\sg[\gamma_\zeta]\,\mathcal Q_{k^\star}(\theta;\zeta).
\end{aligned}
\label{eq:objective}
\end{equation}
The first term retains dense teacher supervision across valid turns using their pre-calibration weights. The second supplements trajectory-relative weighting with supervision on the selected turn's original student response only when $\gamma_\zeta>0$, leaving other turns' pre-calibration weights unchanged.

For this fixed batch, holding recorded $\psi_{k,j}$ and weights fixed during differentiation gives
\begin{equation}
\ell_{\mathrm{pol}}\!\left(\rho_{k,j}(\theta),\sg[\omega_k\psi_{k,j}]\right)
=\sg[\omega_k]\,\ell_{\mathrm{pol}}\!\left(\rho_{k,j}(\theta),\sg[\psi_{k,j}]\right),
\qquad \omega_k\ge0.
\label{eq:weighted-token-loss}
\end{equation}
Thus, turn weighting scales token-level teacher supervision while preserving the directions defined in Section~\ref{sec:opd}. With the original token mask $\mathcal M_k(\zeta)$ and normalization $Z$ unchanged, outcome-based calibration adds a gradient contribution only from the selected candidate turns
\begin{equation}
\nabla_\theta\mathcal L_{\mathrm{OG-OPD}}(\theta)-\nabla_\theta\mathcal L_{\mathrm{base}}(\theta)
=\frac{1}{Z}\sum_{\zeta\in\mathcal D_{\mathrm{pair}}}\sg[\gamma_\zeta]\,\nabla_\theta\mathcal Q_{k^\star}(\theta;\zeta).
\label{eq:calibration-gradient}
\end{equation}
Paired outcomes provide evidence for strengthening supervision at the selected turn, while signed token-level teacher feedback determines the update directions on the original student response. Since $Z$ counts valid original tokens rather than weights, other turns' loss coefficients remain unchanged. Appendix~\ref{app:optimization} provides the derivation and relates observed final task outcomes to training updates.
\par\endgroup

\section{Experiments}
\label{sec:experiments}
\subsection{Experimental Setup}
\label{sec:setup}
\textbf{Benchmarks and Models.} We evaluate OG-OPD on ALFWorld, ScienceWorld, and WebShop, following TCOD \citep{wang2026tcod} for environment configurations, prompt templates, and ALFWorld evaluation collections. All experiments use Qwen3 models \citep{yang2025qwen3}. A Qwen3-32B teacher is paired with a Qwen3-1.7B student on all benchmarks, while ALFWorld and WebShop additionally use a Qwen3-8B-RL teacher with a Qwen3-4B student. Each RL teacher is trained with GiGPO \citep{feng2025gigpo} on its benchmark. We report success rate (SR), task score (Score), and mean interaction rounds (Rounds). Appendices~\ref{app:configurations}--\ref{app:details} detail teacher training and evaluation.

\textbf{Training Setup.} All methods are evaluated at training step~200, with results in the main, ablation, and efficiency tables averaged over three training seeds. Within each benchmark and model setting, all methods share the same training tasks, student initialization, teacher checkpoints, interaction protocol, and evaluation protocol, while retaining their original curriculum and prefix designs. Additional execution details, including action parsing and state replay, are provided in Appendix~\ref{app:implementation}.

\textbf{Baselines.} We compare OG-OPD with five agentic distillation baselines. Vanilla OPD provides dense teacher supervision on the student's own trajectories \citep{agarwal2024onpolicy}. TCOD-F2B progressively increases student rollout depth, whereas TCOD-B2F shortens successful teacher prefixes \citep{wang2026tcod}. Guided-OPD interleaves teacher and student turns while gradually increasing student control \citep{li2026guided}. FutureBridge-OPD identifies large-gap positions, evaluates teacher guidance using teacher preference over student continuations, and adds accepted teacher responses as distillation targets \citep{chen2026futurebridge}. Zero-shot teachers and students serve as references.

\subsection{Main Results and Training Dynamics}
\label{sec:main-results}
\label{sec:training-analysis}
\begin{table}[!t]
\setlength{\abovecaptionskip}{4pt}
\caption{\textbf{Main results with the Qwen3-32B teacher and Qwen3-1.7B student.} Trained methods report mean $\pm$ standard deviation across three training seeds. SR (\%) denotes success rate, with ALFWorld Overall pooling all splits. Score uses a 0--100 scale, and Rounds averages interactions per task. Dark/light green mark the best/second-best means among methods, with the best in bold.}
\label{tab:main}
\centering
\setlength{\tabcolsep}{0pt}
\renewcommand{\arraystretch}{1.2}
\fontsize{9}{10}\selectfont
\begin{tabularx}{\linewidth}{@{}l@{\hspace{1pt}}*{4}{>{\hsize=.94375\hsize\centering\arraybackslash}X}>{\hsize=1.15\hsize\centering\arraybackslash}X*{2}{>{\hsize=.94375\hsize\centering\arraybackslash}X}>{\hsize=1.15\hsize\centering\arraybackslash}X*{2}{>{\hsize=.94375\hsize\centering\arraybackslash}X}>{\hsize=1.15\hsize\centering\arraybackslash}X}
\toprule
& \multicolumn{5}{c}{\textbf{ALFWorld}} & \multicolumn{3}{c}{\textbf{ScienceWorld}} & \multicolumn{3}{c}{\textbf{WebShop}} \\
\cmidrule(lr){2-6}\cmidrule(lr){7-9}\cmidrule(l){10-12}
\textbf{Method} & \shortstack{Seen\\SR $\uparrow$} & \shortstack{Unseen\\SR $\uparrow$} & \shortstack{Hard\\SR $\uparrow$} & \shortstack{Overall\\SR $\uparrow$} & \mbox{Rounds\,$\downarrow$} & SR $\uparrow$ & Score $\uparrow$ & \mbox{Rounds\,$\downarrow$} & SR $\uparrow$ & Score $\uparrow$ & \mbox{Rounds\,$\downarrow$} \\
\midrule
\rowcolor{settinggray}
\multicolumn{12}{c}{Qwen3-32B teacher $\rightarrow$ Qwen3-1.7B student}\\
Student (zero-shot) & \metricvalue{7.1} & \metricvalue{8.2} & \metricvalue{0.0} & \metricvalue{5.3} & \metricvalue{19.8} & \metricvalue{0.2} & \metricvalue{3.8} & \metricvalue{17.0} & \metricvalue{27.0} & \metricvalue{24.2} & \metricvalue{8.1}\\
Teacher (zero-shot) & \metricvalue{37.1} & \metricvalue{35.1} & \metricvalue{9.9} & \metricvalue{28.1} & \metricvalue{18.5} & \metricvalue{28.2} & \metricvalue{53.9} & \metricvalue{12.0} & \metricvalue{48.0} & \metricvalue{47.3} & \metricvalue{4.9}\\
Vanilla OPD & \metricpm{26.7}{3.0} & \metricpm{27.6}{3.3} & \metricpm{4.1}{1.4} & \metricpm{20.1}{2.1} & \metricbest{\metricpm{17.7}{0.4}} & \metricpm{13.0}{1.4} & \metricpm{48.1}{1.2} & \metricpm{12.6}{0.3} & \metricpm{28.0}{2.6} & \metricpm{27.5}{2.6} & \metricpm{6.8}{0.2}\\
TCOD-F2B & \metricpm{23.8}{3.2} & \metricpm{27.9}{3.0} & \metricbest{\metricpm{7.4}{0.8}} & \metricpm{20.2}{2.0} & \metricpm{19.2}{0.3} & \metricpm{13.7}{0.7} & \metricsecond{\metricpm{50.0}{0.5}} & \metricpm{12.5}{0.2} & \metricpm{30.7}{1.5} & \metricpm{25.9}{1.8} & \metricpm{7.7}{0.2}\\
TCOD-B2F & \metricsecond{\metricpm{31.0}{3.6}} & \metricbest{\metricpm{33.6}{3.3}} & \metricpm{5.8}{1.4} & \metricsecond{\metricpm{24.1}{2.3}} & \metricpm{18.5}{0.3} & \metricpm{14.3}{0.5} & \metricpm{48.3}{0.6} & \metricpm{11.9}{0.2} & \metricpm{33.7}{3.1} & \metricpm{33.5}{3.4} & \metricpm{6.1}{0.3}\\
Guided-OPD & \metricpm{26.7}{5.1} & \metricpm{28.4}{4.9} & \metricpm{5.8}{1.7} & \metricpm{20.8}{3.3} & \metricpm{18.4}{0.5} & \metricsecond{\metricpm{15.8}{0.8}} & \metricpm{45.2}{2.0} & \metricpm{12.4}{0.4} & \metricpm{34.3}{9.6} & \metricpm{30.8}{8.9} & \metricpm{5.9}{0.4}\\
FutureBridge-OPD & \metricpm{26.7}{3.3} & \metricpm{29.4}{3.0} & \metricsecond{\metricpm{6.6}{0.8}} & \metricpm{21.4}{2.1} & \metricpm{18.1}{0.3} & \metricpm{12.8}{0.7} & \metricpm{42.6}{0.5} & \metricsecond{\metricpm{11.7}{0.2}} & \metricsecond{\metricpm{38.7}{1.5}} & \metricsecond{\metricpm{38.1}{1.6}} & \metricsecond{\metricpm{5.7}{0.2}}\\
\textbf{OG-OPD} & \metricbest{\metricpm{31.7}{2.1}} & \metricsecond{\metricpm{32.3}{2.3}} & \metricsecond{\metricpm{6.6}{0.8}} & \metricbest{\metricpm{24.2}{1.5}} & \metricsecond{\metricpm{18.0}{0.2}} & \metricbest{\metricpm{16.6}{0.6}} & \metricbest{\metricpm{52.0}{0.7}} & \metricbest{\metricpm{11.6}{0.2}} & \metricbest{\metricpm{45.7}{1.2}} & \metricbest{\metricpm{44.3}{1.5}} & \metricbest{\metricpm{5.6}{0.2}}\\
\bottomrule
\end{tabularx}
\end{table}

\begin{table}[!t]
\vspace*{6pt}
\captionsetup{position=top,skip=4pt}
\caption{\textbf{Main results with the Qwen3-8B-RL teacher and Qwen3-4B student.} Each teacher is trained with GiGPO on its benchmark. Results for trained methods are mean $\pm$ standard deviation across three training seeds. SR (\%), Score (0--100), and Rounds follow Table~\ref{tab:main}. Dark/light green mark the best/second-best means among compared methods, with the best in bold.}
\label{tab:rl}
\centering
\setlength{\tabcolsep}{1pt}
\renewcommand{\arraystretch}{1.1}
\fontsize{9}{10}\selectfont
\begin{tabularx}{\linewidth}{@{}l*{8}{>{\centering\arraybackslash}X}}
\toprule
& \multicolumn{5}{c}{\textbf{ALFWorld}} & \multicolumn{3}{c}{\textbf{WebShop}} \\
\cmidrule(lr){2-6}\cmidrule(l){7-9}
\textbf{Method} & \shortstack{Seen\\SR $\uparrow$} & \shortstack{Unseen\\SR $\uparrow$} & \shortstack{Hard\\SR $\uparrow$} & \shortstack{Overall\\SR $\uparrow$} & \mbox{Rounds\,$\downarrow$} & SR $\uparrow$ & Score $\uparrow$ & \mbox{Rounds\,$\downarrow$} \\
\midrule
\rowcolor{settinggray}
\multicolumn{9}{c}{Qwen3-8B-RL teacher $\rightarrow$ Qwen3-4B student}\\
Student (zero-shot) & \metricvalue{32.1} & \metricvalue{26.9} & \metricvalue{9.9} & \metricvalue{23.5} & \metricvalue{17.5} & \metricvalue{27.0} & \metricvalue{26.5} & \metricvalue{8.0}\\
Teacher (zero-shot) & \metricvalue{81.4} & \metricvalue{79.1} & \metricvalue{33.9} & \metricvalue{66.1} & \metricvalue{10.3} & \metricvalue{54.0} & \metricvalue{56.4} & \metricvalue{4.4}\\
Vanilla OPD & \metricpm{71.4}{3.7} & \metricpm{67.4}{3.5} & \metricpm{20.9}{2.4} & \metricpm{54.6}{2.7} & \metricpm{11.3}{0.3} & \metricpm{51.7}{1.5} & \metricpm{55.2}{2.2} & \metricpm{4.2}{0.1}\\
TCOD-F2B & \metricpm{77.9}{1.4} & \metricpm{71.9}{1.7} & \metricpm{28.1}{1.4} & \metricpm{60.6}{1.2} & \metricpm{11.5}{0.2} & \metricsecond{\metricpm{55.3}{2.3}} & \metricsecond{\metricpm{57.8}{2.8}} & \metricsecond{\metricpm{4.0}{0.1}}\\
TCOD-B2F & \metricpm{77.1}{2.1} & \metricpm{75.6}{1.9} & \metricbest{\metricpm{31.4}{1.4}} & \metricpm{62.6}{1.6} & \metricpm{11.3}{0.2} & \metricpm{54.7}{1.2} & \metricpm{56.6}{1.5} & \metricpm{4.1}{0.1}\\
Guided-OPD & \metricpm{74.3}{3.8} & \metricpm{73.4}{3.5} & \metricpm{19.3}{2.5} & \metricpm{57.1}{2.8} & \metricsecond{\metricpm{11.2}{0.3}} & \metricpm{52.3}{4.6} & \metricpm{53.7}{3.6} & \metricpm{4.3}{0.2}\\
FutureBridge-OPD & \metricsecond{\metricpm{78.6}{2.9}} & \metricbest{\metricpm{77.1}{2.8}} & \metricpm{28.4}{1.9} & \metricsecond{\metricpm{62.7}{2.1}} & \metricpm{11.5}{0.2} & \metricpm{54.7}{0.6} & \metricpm{56.9}{1.1} & \metricpm{4.1}{0.1}\\
\textbf{OG-OPD} & \metricbest{\metricpm{82.1}{1.9}} & \metricsecond{\metricpm{76.4}{1.9}} & \metricsecond{\metricpm{30.0}{1.3}} & \metricbest{\metricpm{64.2}{1.4}} & \metricbest{\metricpm{11.1}{0.2}} & \metricbest{\metricpm{56.3}{1.2}} & \metricbest{\metricpm{57.9}{1.3}} & \metricbest{\metricpm{3.9}{0.1}}\\
\bottomrule
\end{tabularx}
\end{table}

\begin{figure}[!t]
\centering
\includegraphics[width=\linewidth]{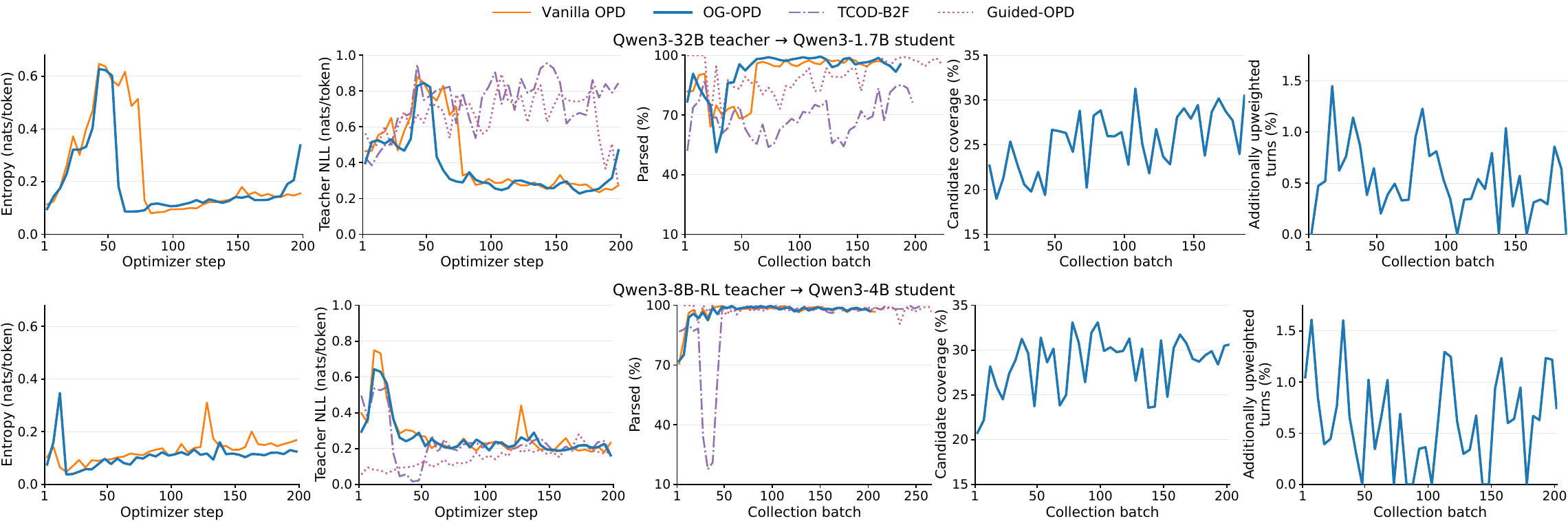}%
\caption{\textbf{Training dynamics on WebShop.} The top row shows the Qwen3-32B teacher with the Qwen3-1.7B student. The bottom row shows the Qwen3-8B-RL teacher with the Qwen3-4B student.}
\label{fig:dynamics}
\par\vspace{-1pt}
\begingroup
\captionsetup{type=table}
\caption{\textbf{Ablation results.} The first group tests outcome-based calibration by removing it or directly upweighting candidate turns without verification. The second retains calibration and teacher response evaluation but replaces trajectory-relative weighting with uniform unit weights or gap-based candidate selection with random turn sampling. Trained methods report mean $\pm$ standard deviation across three training seeds. Bold marks each column's best mean among trained methods.}
\label{tab:ablation}
\centering
\setlength{\tabcolsep}{3pt}
\renewcommand{\arraystretch}{1.0}
\fontsize{9}{10}\selectfont
\definecolor{ablationcalibration}{HTML}{DCE9D4}
\definecolor{ablationselection}{HTML}{D9EAE8}
\begin{tabularx}{\linewidth}{@{}l*{6}{>{\centering\arraybackslash}X}@{}}
\toprule
& \multicolumn{3}{c}{\textbf{ScienceWorld}} & \multicolumn{3}{c}{\textbf{WebShop}}\\
\cmidrule(lr){2-4}\cmidrule(l){5-7}
\textbf{Method} & SR $\uparrow$ & Score $\uparrow$ & Rounds $\downarrow$ & SR $\uparrow$ & Score $\uparrow$ & Rounds $\downarrow$\\
\midrule
\rowcolor{settinggray}
\multicolumn{7}{c}{Qwen3-32B teacher $\rightarrow$ Qwen3-1.7B student}\\
Student (zero-shot) & \metricvalue{0.2} & \metricvalue{3.8} & \metricvalue{17.0} & \metricvalue{27.0} & \metricvalue{24.2} & \metricvalue{8.1}\\
Teacher (zero-shot) & \metricvalue{28.2} & \metricvalue{53.9} & \metricvalue{12.0} & \metricvalue{48.0} & \metricvalue{47.3} & \metricvalue{4.9}\\
Vanilla OPD & \metricpm{13.0}{1.4} & \metricpm{48.1}{1.2} & \metricpm{12.6}{0.3} & \metricpm{28.0}{2.6} & \metricpm{27.5}{2.6} & \metricpm{6.8}{0.2}\\
\midrule
\rowcolor{ablationcalibration}
\multicolumn{7}{c}{\textbf{Outcome-Based Calibration}}\\
w/o outcome-based calibration & \metricpm{3.8}{1.1} & \metricpm{38.5}{1.9} & \metricpm{12.5}{0.3} & \metricpm{32.0}{2.6} & \metricpm{30.5}{2.5} & \metricpm{6.4}{0.2}\\
Upweighting without verification & \metricpm{14.9}{0.9} & \metricpm{49.9}{1.1} & \metricpm{12.3}{0.2} & \metricpm{35.0}{2.0} & \metricpm{34.5}{2.1} & \metricpm{6.1}{0.2}\\
\midrule
\rowcolor{ablationselection}
\multicolumn{7}{c}{\textbf{Weighting and Candidate Selection}}\\
w/o trajectory-relative weighting & \metricpm{5.5}{1.3} & \metricpm{38.1}{2.1} & \metricpm{12.9}{0.3} & \metricpm{33.3}{2.5} & \metricpm{31.9}{2.8} & \metricpm{6.1}{0.2}\\
Random turn sampling & \metricpm{15.1}{0.8} & \metricpm{51.5}{0.9} & \metricpm{12.1}{0.2} & \metricpm{37.7}{1.5} & \metricpm{35.6}{2.0} & \metricpm{5.9}{0.2}\\
\midrule
\textbf{OG-OPD} & \textbf{\metricpm{16.6}{0.6}} & \textbf{\metricpm{52.0}{0.7}} & \textbf{\metricpm{11.6}{0.2}} & \textbf{\metricpm{45.7}{1.2}} & \textbf{\metricpm{44.3}{1.5}} & \textbf{\metricpm{5.6}{0.2}}\\
\bottomrule
\end{tabularx}
\endgroup
\end{figure}

OG-OPD achieves the highest SR among the compared methods in every evaluated setting, with Overall SR reported for ALFWorld, as shown in Tables~\ref{tab:main} and~\ref{tab:rl}. Relative to vanilla OPD, the gains range from 3.6 to 17.7 percentage points. With the Qwen3-32B teacher and Qwen3-1.7B student, OG-OPD reaches 45.7\% SR on WebShop, compared with 38.7\% for FutureBridge-OPD, the strongest competing baseline. The advantage also holds with the task-trained Qwen3-8B-RL teacher and Qwen3-4B student. On ALFWorld, OG-OPD achieves 64.2\% Overall SR, surpassing FutureBridge-OPD at 62.7\%. Both students also outperform vanilla OPD on Seen, Unseen, and Hard tasks, showing consistent gains in unfamiliar environments and on difficult tasks.

With the Qwen3-32B teacher and Qwen3-1.7B student, OG-OPD improves SR over vanilla OPD by 17.7 percentage points on WebShop, compared with 3.6 points on ScienceWorld. This contrast is consistent with the empirical analysis in Section~\ref{sec:diagnosis}, which shows that teacher harm is more frequent than teacher rescue within WebShop's large-gap group, while the reverse holds on ScienceWorld. The larger improvement on WebShop, together with this empirical finding, supports the motivation for checking whether teacher guidance benefits the current student before strengthening supervision.

The gains extend beyond the binary success criterion used for outcome-based calibration. Across all evaluated ScienceWorld and WebShop settings, OG-OPD achieves the highest Score and the lowest Rounds among the compared methods. For Qwen3-1.7B, ScienceWorld Score increases from 48.1 with vanilla OPD to 52.0, while Rounds decreases from 12.6 to 11.6. On WebShop, Score increases from 27.5 to 44.3, while Rounds decreases from 6.8 to 5.6. The Qwen3-4B student on WebShop shows the same pattern. Higher task success accompanies higher task scores and shorter interactions, even though neither Score nor Rounds is used to trigger additional upweighting.

On WebShop, OG-OPD shows earlier improvements in teacher likelihood and action parseability than vanilla OPD. With the Qwen3-32B teacher and Qwen3-1.7B student, Teacher NLL falls earlier and Parsed recovers sooner, as shown in Figure~\ref{fig:dynamics}. These improvements accompany selective additional upweighting. Across both teacher--student configurations, candidate turns account for roughly one quarter of eligible turns, whereas fewer than 1\% of recorded turns receive additional weight when pooled over training. Thus, earlier response-level improvements coexist with additional upweighting of only a small subset of turns, while dense teacher supervision is retained across the student's own trajectories. These gains come at a per-step training cost of 1.42$\times$ vanilla OPD on ScienceWorld and 1.48$\times$ on WebShop with the Qwen3-1.7B student, representing a practical tradeoff between training cost and task performance. Appendix~\ref{app:efficiency} provides training time details.

\subsection{Ablation Study}
\label{sec:ablations}

We evaluate outcome-based calibration, trajectory-relative weighting, and candidate selection using the Qwen3-32B teacher and Qwen3-1.7B student, as shown in Table~\ref{tab:ablation}. We first test whether simpler weighting strategies can replace outcome-based calibration. The \emph{w/o outcome-based calibration} variant retains the trajectory-relative weights without further adjustment, while \emph{Upweighting without verification} directly increases the weight of the earliest gap-based candidate turn. Neither variant generates teacher responses or performs paired student continuations. Direct upweighting improves SR from 3.8\% to 14.9\% on ScienceWorld and from 32.0\% to 35.0\% on WebShop, but still falls short of OG-OPD at 16.6\% and 45.7\%, respectively. Increasing candidate turn weights improves SR on both benchmarks, but does not match the gains from the full outcome-based calibration procedure.

With outcome-based calibration retained, we next examine the role of trajectory-relative weighting. The \emph{w/o trajectory-relative weighting} variant replaces the pre-calibration weights with uniform unit weights while keeping candidate selection, teacher response evaluation, and calibration unchanged. Relative to OG-OPD, SR drops by 11.1 percentage points on ScienceWorld and 12.4 points on WebShop. These results show that trajectory-relative weighting and outcome-based calibration are complementary, as removing either component substantially reduces performance.

Finally, we examine the effect of candidate turn selection. The \emph{Random turn sampling} variant randomly selects eligible turns without replacement from trajectories containing at least one candidate turn, with teacher response evaluation and outcome-based calibration unchanged. Relative to OG-OPD, SR drops by 1.5 percentage points on ScienceWorld and 8.0 points on WebShop. Gap-based candidate selection thus helps outcome-based calibration focus on more informative turns.

\section{Related Work}
\label{sec:related}
\textbf{On-Policy Distillation for Multi-Turn Agents.} Learning from student-generated responses addresses the training--inference distribution mismatch in autoregressive distillation \citep{agarwal2024onpolicy}. Subsequent research spans objective design and rollout construction. Objective-level studies investigate reverse KL distillation \citep{gu2024minillm}, adaptive intermediate distributions \citep{jang2026stable}, and entropy-dependent objectives \citep{jin2026entropy}. Complementary work combines skew KL objectives with adaptive reuse of student-generated outputs \citep{ko2024distillm}. In multi-turn environments, temporal curricula regulate student rollout depth \citep{wang2026tcod}, while turn-level guidance schedules teacher participation within a rollout \citep{li2026guided}. Prefix replay organizes student continuations using previously collected teacher trajectories \citep{liao2026prefix}. These approaches shape the states visited during training and the distributions used as learning targets.

\textbf{Selective Supervision and Reweighting.} In chain-of-thought distillation, learned token weights emphasize key reasoning tokens \citep{feng2024keypoint}. For OPD, supervision strength is adjusted across tokens or steps using entropy and teacher--student token-level distributional gaps \citep{xu2026tip}, or final task outcomes and perplexity \citep{zheng2026scope}. A trajectory-relative weighting rule interprets these gaps as a reliability signal and attenuates supervision as their magnitude increases relative to a reference \citep{zhong2026sod}. Although its overall objective includes trajectory rewards, the distillation weights depend on relative gap magnitudes rather than the task outcomes induced by teacher actions. The weighting rule therefore does not distinguish downweighted turns where teacher guidance would help the current student from those where it would not.

\textbf{Downstream Evaluation and Teacher Guidance.} In reasoning tasks, sampled continuations provide automatic step-level supervision \citep{wang2024mathshepherd} and Monte Carlo estimates for credit assignment \citep{kazemnejad2025vineppo}. Policy optimization also uses discounted future KL to shape token-level advantages \citep{ma2026fipo} and repeated-state action groups for step-level credit assignment \citep{feng2025gigpo}. In agent distillation, \citet{chen2026futurebridge} evaluate teacher responses through increases in the proportion of teacher-preferred tokens in paired student continuations, retaining accepted teacher responses as additional distillation targets. OG-OPD likewise evaluates teacher guidance through paired student continuations, but differs in both the validation signal and how it enters training. It uses final task outcomes from paired student continuations to decide whether to strengthen teacher supervision at the selected turn of the student's original trajectory, rather than adding accepted teacher responses as additional distillation targets for student policy optimization.

\section{Conclusion}
For multi-turn autonomous agents, teacher--student token-level distributional gaps are not enough to determine whether teacher guidance benefits the current student. Outcome-Guided On-Policy Distillation (OG-OPD) addresses this supervision--benefit mismatch by calibrating trajectory-relative weights with final task outcomes from paired student continuations while training on the student's original trajectories. Across three benchmarks and two teacher--student pairings, OG-OPD achieves the highest mean task success rate among the compared methods in every evaluated setting.

\subsection*{AI use statement}
Generative AI tools were used to polish the paper and assist with experimental code development. They were not used for any other tasks requiring disclosure. All AI-assisted work was reviewed by the authors. AI-assisted code was reviewed and tested for correctness, and language edits were checked to ensure they preserved the intended technical content. We take responsibility for the final content of this work, including text, claims, or artifacts produced with the aid of generative AI.

\subsection*{Ethics statement}
We evaluate language agents in benchmark environments, without live purchases or physical deployments. Improved benchmark performance does not establish safety in real-world settings, where erroneous actions may cause consequences not captured by these environments. Deployment therefore requires separate safety evaluations and appropriate safeguards. All datasets, environments, and models remain subject to their respective licenses and terms of use.

\subsection*{Reproducibility statement}
We provide the source code in the supplementary material accompanying this submission. Section~\ref{sec:method} presents the empirical analysis and the OG-OPD algorithm, with optimization details, diagnostic protocols, and implementation procedures in Appendices~\ref{app:optimization}--\ref{app:implementation}. Appendix~\ref{app:configurations} specifies the model configurations, and Appendix~\ref{app:details} describes the evaluation metrics and aggregation across training seeds. The training cost measurement protocol is provided in Appendix~\ref{app:efficiency}. Together, these materials document the method and experimental procedures needed to reproduce the reported results.

\bibliography{cgd_references}
\bibliographystyle{conference}

\clearpage
\appendix
\setlength{\parskip}{3pt plus 1pt minus 1pt}
\setlength{\textfloatsep}{9pt plus 1pt minus 1pt}
\setlength{\floatsep}{4pt plus 1pt minus 1pt}
\setlength{\intextsep}{9pt plus 1pt minus 1pt}
\AddToHook{cmd/normalsize/after}[appendix-spacing]{%
  \setlength{\abovedisplayskip}{3.5pt plus 1pt minus 1pt}%
  \setlength{\belowdisplayskip}{3.5pt plus 1pt minus 1pt}%
  \setlength{\abovedisplayshortskip}{1pt plus 1pt}%
  \setlength{\belowdisplayshortskip}{3pt plus 1pt minus 1pt}%
}
\normalsize
\section{Optimization details}
\label{app:optimization}
The paired comparison in Equation~\ref{eq:pair} assumes successful replay to the same state and history before the selected response. With independent environment instances, a common continuation policy, and equal remaining horizons, the comparison conditions are
\begin{equation}
\begin{aligned}
(s_{k^\star}^{S},c_{k^\star}^{S})&=(s_{k^\star}^{T},c_{k^\star}^{T})=(s_{k^\star},c_{k^\star}),\\
p_{\mathrm{cont}}^{S}&=p_{\mathrm{cont}}^{T}=p_\circ,
\qquad B_{\mathrm{rem}}^{S}=B_{\mathrm{rem}}^{T}.
\end{aligned}
\label{eq:matched-state}
\end{equation}
As in Section~\ref{sec:calibration}, $T$ identifies the use of the teacher response, which replaces the student's original response in the assistant history. Both continuations share decoding settings and seed schedules. Replay or runtime errors exclude a trajectory from $\mathcal D_{\mathrm{pair}}$. The positive part of the final task outcome difference, $\pos{\Delta_\zeta}$, equals the indicator $\mathsf g_\zeta$ for additional teacher supervision in Equation~\ref{eq:gate}.

For a fixed batch $\mathcal D$ of the student's own trajectories, vanilla OPD and OG-OPD apply the same clipped policy surrogate to the same valid original response tokens. Only the turn weights differ. The valid token set and normalization are
\begin{equation}
\mathcal M_{\mathrm{orig}}(\zeta)=\{(k,j):j\in\mathcal M_k(\zeta)\},
\qquad
Z=\sum_{\zeta\in\mathcal D}|\mathcal M_{\mathrm{orig}}(\zeta)|,
\label{eq:original-tokens}
\end{equation}
where $\mathcal M_k(\zeta)$ excludes prompts, environment observations, padding, and positions without valid teacher scores. The following identities concern batches with $Z>0$. For importance ratio $\rho$ and fixed signed coefficient $a$ representing token-level teacher supervision, the shared surrogate is
\begin{equation}
\ell_{\mathrm{ratio}}(\rho,a)=\max\!\left\{-\rho a,-\operatorname{clip}(\rho,1-\varepsilon_-,1+\varepsilon_+)a\right\},
\label{eq:clip-base}
\end{equation}
\begin{equation}
\ell_{\mathrm{pol}}(\rho,a)=
\begin{cases}
\min\{\ell_{\mathrm{ratio}}(\rho,a),-\kappa a\},&a<0,\\
\ell_{\mathrm{ratio}}(\rho,a),&a\ge0.
\end{cases}
\label{eq:clip-loss}
\end{equation}
Here $\varepsilon_-,\varepsilon_+>0$ are clipping widths and $\kappa>1$ bounds the surrogate for $a<0$. Token-level teacher supervision retains the signed coefficients $\psi_{k,j}$ defined in Section~\ref{sec:opd}. Recorded probabilities, turn weights, and final task outcome indicators are held fixed during differentiation. In implementation, the surrogate uses the numerically bounded ratio
\begin{equation}
\widehat\rho_{k,j}(\theta)=\exp\!\left(\operatorname{clip}\!\left(
\log p_\theta(z_{k,j}\mid v_{k,j})-\log p_\circ(z_{k,j}\mid v_{k,j}),-C,C\right)\right),
\label{eq:stable-ratio}
\end{equation}
where $C=20$. The bounded ratio equals Equation~\ref{eq:signal} when clipping is inactive. Equation~\ref{eq:homogeneity} holds for either ratio. The following identities apply to the fixed-coefficient policy surrogate, not the exact reverse-KL objective at arbitrary student parameters.

All valid original tokens at a turn share its weight, applied through the recorded advantage
\begin{equation}
a_{k,j}^{\mathrm{OG-OPD}}=\sg\!\left[\omega_k\psi_{k,j}\right].
\label{eq:advantage}
\end{equation}
The clipped loss is positively homogeneous in its signed argument. For fixed $\omega\ge0$,
\begin{equation}
\ell_{\mathrm{pol}}(\rho,\omega a)=\omega\ell_{\mathrm{pol}}(\rho,a).
\label{eq:homogeneity}
\end{equation}
Multiplication by a positive fixed weight preserves the sign of $a$ and the ordering of the arguments to each minimum and maximum; both sides vanish for $\omega=0$. On this fixed batch, weighting the recorded advantage is therefore equivalent to weighting the original turn loss $\mathcal Q_k$. Substituting Equation~\ref{eq:weight} yields the decomposition in Equation~\ref{eq:objective}.

Under Equation~\ref{eq:weight}, $\gamma_\zeta=0$ when $\mathsf g_\zeta=0$ or $\beta_{k^\star}\ge\beta_{\uparrow}$. With no additional weight, the objective reduces to $\mathcal L_{\mathrm{base}}$. Unit pre-calibration weights further recover Equation~\ref{eq:opd-objective}. For fixed recorded coefficients, positive turn weights rescale each token's loss gradient without reversing its direction. Reweighting across turns can nevertheless change the direction of the batch gradient.

\paragraph{Gradient decomposition and weight bounds.}
For the fixed batch and recorded weights above, differentiation can be taken inside the finite token sums:
\begin{equation}
\nabla_\theta\mathcal L_{\mathrm{OG-OPD}}
=\frac{1}{Z}\sum_{\zeta\in\mathcal D}\sum_{k\in\mathcal V(\zeta)}\omega_k
\sum_{j\in\mathcal M_k(\zeta)}\nabla_\theta
\ell_{\mathrm{pol}}\!\left(\rho_{k,j}(\theta),\sg[\psi_{k,j}]\right).
\label{eq:token-gradient-expansion}
\end{equation}
Splitting $\omega_k$ into its base and additional terms recovers Equation~\ref{eq:calibration-gradient}, with no gradient through the sampled final task outcomes. Wherever the derivatives exist, the triangle inequality gives
\begin{equation}
\left\|\nabla_\theta\mathcal L_{\mathrm{OG-OPD}}-\nabla_\theta\mathcal L_{\mathrm{base}}\right\|
\le\frac{1}{Z}\sum_{\zeta\in\mathcal D_{\mathrm{pair}}}
\gamma_\zeta\left\|\nabla_\theta\mathcal Q_{k^\star}(\theta;\zeta)\right\|.
\label{eq:calibration-gradient-bound}
\end{equation}
This expression bounds the additional loss gradient in terms of the selected turn gradients and their weights. A small number of upweighted turns alone does not imply a small gradient change, and the bound does not establish an improvement in task success after optimization.

The floor rule provides explicit bounds on the coefficients. Since $0<\beta_k\le\beta_{\max}$ on valid turns and $\mathsf g_\zeta\in\{0,1\}$,
\begin{equation}
0\le\gamma_\zeta\le\beta_{\uparrow},\qquad
\beta_k\le\omega_k\le\max\{\beta_{\max},\beta_{\uparrow}\}.
\label{eq:calibrated-weight-bounds}
\end{equation}
These are bounds on supervision coefficients, not on gradient norms. In particular, no additional normalization by the sum of weights is introduced, so the original valid token denominator $Z$ is preserved.

\paragraph{Interpretation of outcome-based calibration.}
For a restored state $s$, history $c$, response $z$, and frozen current student, define the probability of a successful final task outcome by
\begin{equation}
V_{p_\circ}(s,c,z)=\mathbb E_\varsigma\!\left[\mathsf S\!\left(\mathcal U(s,c,z;p_\circ,\varsigma)\right)\right].
\label{eq:completion-probability}
\end{equation}
For fixed $(s,c,z^S,z^T,p_\circ)$, correctly restored states and the intended sampling marginals give
\begin{equation}
\mathbb E_\varsigma\!\left[\mathsf S(\tilde\zeta^T)-\mathsf S(\tilde\zeta^S)\right]
=V_{p_\circ}(s,c,z^T)-V_{p_\circ}(s,c,z^S).
\label{eq:expected-outcome-difference}
\end{equation}
The shared seed schedule couples the final task outcomes without changing this identity. Under these fixed conditions,
\begin{equation}
\begin{aligned}
\mathbb E_\varsigma[\Delta_\zeta]
&=\Pr_\varsigma(\Delta_\zeta=1)-\Pr_\varsigma(\Delta_\zeta=-1),\\
\mathbb E_\varsigma[\mathsf g_\zeta]
&=\Pr_\varsigma(\Delta_\zeta=1).
\end{aligned}
\label{eq:rescue-versus-net-benefit}
\end{equation}
The first expectation measures net execution benefit; the second measures teacher rescue probability under paired sampling. OG-OPD uses one observed teacher rescue to condition additional weight, rather than estimating net expected benefit from repeated trials. These execution statistics do not establish the effect of an individual training update.

For a selected turn with fixed pre-calibration weight, let $h_\zeta=[\beta_{\uparrow}-\beta_{k^\star}]_+$ and $p_+=\Pr_\varsigma(\Delta_\zeta=1)$ under the same fixed comparison conditions. Since $\gamma_\zeta=h_\zeta\mathsf g_\zeta$, its first two moments are
\begin{equation}
\mathbb E_\varsigma[\gamma_\zeta]=h_\zeta p_+,\qquad
\operatorname{Var}_\varsigma(\gamma_\zeta)=h_\zeta^2p_+(1-p_+).
\label{eq:additional-weight-moments}
\end{equation}
The expected additional weight therefore depends on both the probability of an observed rescue and the distance to the weight floor. When $h_\zeta=0$, even a positive paired outcome leaves the weight unchanged. For $h_\zeta>0$, the probability of an actual increase equals $p_+$ under these fixed conditions.

\section{Empirical analysis details}
\label{app:diagnostics}
Section~\ref{sec:diagnosis} introduces the empirical analysis with a fixed Qwen3-1.7B student and Qwen3-32B teacher on ScienceWorld and WebShop. Continuations $S$ and $TS$ use the current student after the student's original action and the teacher action, respectively; $TT$ uses the teacher after the same teacher action to compare what the teacher and current student can complete. Eligible positions within the student's own trajectories are sampled uniformly without replacement, independently of OG-OPD's candidate selection rule. Under a bounded proposal budget, we retain the first replay-valid sampled position with a teacher response whose parsed action is legal and differs from the student's original action. The empirical analysis therefore includes only decisions with valid recorded scores, successful replay, and valid teacher responses.

Each retained decision is evaluated in five matched trials. Let $\tilde\zeta_m^b$ be continuation $b$ in trial $m$. With $M$ valid trials, we compute each continuation's success frequency and how often the paired student continuations differ in final task outcome:
\begin{equation}
f_b=\frac{1}{M}\sum_{m=1}^{M}\mathsf S(\tilde\zeta_m^b),\quad b\in\{S,TS,TT\},\qquad
q_{\mathrm{diff}}=\frac{1}{M}\sum_{m=1}^{M}|\mathsf S(\tilde\zeta_m^{TS})-\mathsf S(\tilde\zeta_m^S)|.
\label{eq:diagnostic-frequencies}
\end{equation}
We assign outcome categories using lower and upper success frequency thresholds $\alpha_-<\alpha_+$:
\begin{equation}
\begin{aligned}
\text{benign gaps}:&\quad f_S\ge\alpha_+,\ f_{TS}\ge\alpha_+,\\
\text{teacher rescue}:&\quad f_S\le\alpha_-,\ f_{TS}\ge\alpha_+,\\
\text{teacher harm}:&\quad f_S\ge\alpha_+,\ f_{TS}\le\alpha_-,\\
\text{teacher-only completion}:&\quad f_{TT}\ge\alpha_+,\ f_{TS}\le\alpha_-,\\
\text{outcome-critical}:&\quad q_{\mathrm{diff}}\ge\alpha_+.
\end{aligned}
\label{eq:diagnostic-labels}
\end{equation}
Here $M=5$, $\alpha_-=1/5$, and $\alpha_+=3/5$, corresponding to at most one and at least three successes. Benign gaps use each action's success frequency, whereas outcome-critical decisions require different final task outcomes in at least three matched pairs. Categories can overlap, including teacher harm and teacher-only completion. Model parameters remain fixed, and all three continuations use matched seed schedules, decoding settings, and remaining interaction budgets. Teacher responses replace both the executable action and the assistant response retained in history.

The retained set contains 329 of 330 selected ScienceWorld decisions after excluding one replay failure, and all 274 selected WebShop decisions. We divide gap magnitudes into five groups using empirical quantiles of $\chi_k$ in Equation~\ref{eq:response-gap}; large and small gaps refer to the highest and lowest groups. Boundaries are computed separately over all eligible positions in each benchmark, including unselected positions. In ScienceWorld, the large- and small-gap groups contain 172 and 37 decisions, respectively; the corresponding counts for WebShop are 73 and 33. Empirical analysis quantiles give equal weight to each eligible position, whereas the training quantiles in Equation~\ref{eq:thresholds} balance trajectories. Decisions with completed trials are assigned to these existing groups, yielding unequal group sizes.

\paragraph{Counts and cross-benchmark aggregation.}
Table~\ref{tab:diagnostic-counts} gives the counts behind Figure~\ref{fig:outcomes}. For outcome category $j$, let $I_j(d)$ equal one when decision $d$ meets its criterion and zero otherwise. Let $\mathcal D_{j,e}$ contain the decisions in the corresponding gap group of benchmark $e$. We divide the number of decisions in the category by the group size, then average the two benchmark percentages with equal weight:
\begin{equation}
r_{j,e}=\frac{100}{|\mathcal D_{j,e}|}\sum_{d\in\mathcal D_{j,e}}I_j(d),\qquad
\bar r_j=\frac{r_{j,\mathrm{ScienceWorld}}+r_{j,\mathrm{WebShop}}}{2}.
\label{eq:diagnostic-aggregation}
\end{equation}
The calculation applies to all five categories, with percentages averaged before rounding. Repeated trials determine category membership; each decision contributes once to every category it meets. Rates use the number of retained decisions in the corresponding gap group as their denominator.

\paragraph{Confidence intervals.}
Within each benchmark and gap group, we resample retained decisions with replacement 2,000 times and take the 2.5th and 97.5th percentiles to obtain the 95\% confidence intervals in Figure~\ref{fig:outcomes}. Each decision is classified using its five matched trials before resampling. The zero observed count for small-gap outcome-critical decisions on ScienceWorld produces a degenerate bootstrap interval, not evidence of a zero population proportion.

\begin{table}[htbp]
\centering
\caption{\textbf{Counts underlying the empirical analysis.} Each benchmark entry gives the number of decisions in an outcome category divided by the corresponding gap group size, followed by the percentage in parentheses. Categories follow Equation~\ref{eq:diagnostic-labels}. The average gives equal weight to the two benchmarks.}
\label{tab:diagnostic-counts}
\small
\begin{tabular*}{\linewidth}{@{\extracolsep{\fill}}lcccc}
\toprule
Outcome category & Gap group & ScienceWorld & WebShop & Average\\
\midrule
Benign gaps & Large & 2/172 (1.2\%) & 7/73 (9.6\%) & 5.4\%\\
Teacher rescue & Large & 9/172 (5.2\%) & 9/73 (12.3\%) & 8.8\%\\
Teacher harm & Large & 4/172 (2.3\%) & 14/73 (19.2\%) & 10.8\%\\
Teacher-only completion & Large & 26/172 (15.1\%) & 19/73 (26.0\%) & 20.6\%\\
Outcome-critical & Small & 0/37 (0.0\%) & 5/33 (15.2\%) & 7.6\%\\
\bottomrule
\end{tabular*}
\end{table}

\subsection{Ranking by Gap Magnitude}
\label{app:selection}
\begin{figure}[ht]
\centering
\includegraphics[width=0.90\linewidth]{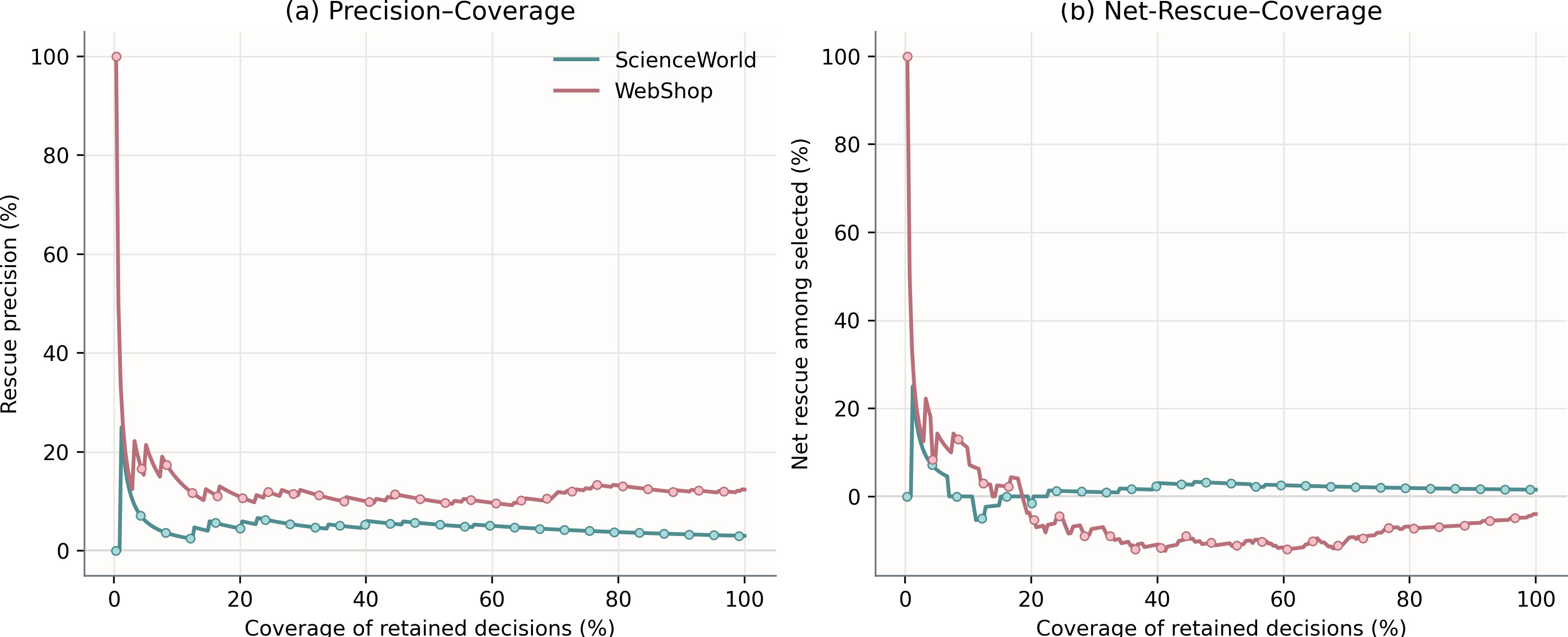}
\caption{\textbf{Final task outcomes under gap magnitude ranking.} Decisions are ranked by decreasing gap magnitude. Rescue precision (left) is the teacher rescue rate; net rescue (right) subtracts the teacher harm rate.}
\label{fig:selection}
\end{figure}
We further examine whether larger gaps preferentially identify decisions where teacher guidance benefits the current student. For Figure~\ref{fig:selection}, let $\mathcal S_k$ contain the top $k$ of $n$ retained decisions ranked by decreasing gap magnitude. With teacher rescue and teacher harm indicators $R_j$ and $H_j$ from Equation~\ref{eq:diagnostic-labels}, the plotted measures are
\begin{equation}
\operatorname{coverage}(k)=\frac{k}{n},\qquad
\operatorname{precision}(k)=\frac{\sum_{j\in\mathcal S_k}R_j}{k},\qquad
\operatorname{net}(k)=\frac{\sum_{j\in\mathcal S_k}(R_j-H_j)}{k}.
\label{eq:ranking-metrics}
\end{equation}
Ties are broken deterministically by task identity. WebShop's net rescue is negative over much of the coverage range, while ScienceWorld's stays near zero or slightly positive. These curves compare the frequencies of the decision categories in Equation~\ref{eq:diagnostic-labels}, not the expected outcome difference in Equation~\ref{eq:expected-outcome-difference}. They describe the retained decisions under the frozen current student, with small selected sets at low coverage.

\section{Execution contract}
\label{app:implementation}
ALFWorld, ScienceWorld, and WebShop responses are parsed into their respective household commands, scientific actions, and search or click operations. The environment receives the parsed action, while the complete assistant response remains in the interaction history. For the table comparisons, methods within each benchmark share evaluation prompts, response token budgets, history retention, action parsing, interaction horizons, termination handling, and score extraction. Action parsing follows each benchmark's shared protocol, without additional repair rules for individual methods.

Teacher and student log probabilities are evaluated on the same valid original response tokens, including generated reasoning and action text but excluding prompts, observations, and padding. Gap magnitudes from Equation~\ref{eq:response-gap} determine trajectory-relative weighting and candidate selection, while the signed coefficients in Equation~\ref{eq:signal} determine token-level teacher supervision. Missing or nonfinite scores are excluded. Turns with empty valid token masks have no gap or weight and are excluded from the loss and threshold populations; trajectories with no valid turns contribute neither supervision nor candidates. A first valid turn $k_0>0$ supplies the weighting reference but has no valid change from the preceding turn and cannot use the candidate exception reserved for $k=0$.

For the candidate thresholds, let $\mathcal H_\zeta^f$ contain eligible positions for statistic $f\in\{\nu,\upsilon\}$ and let $\mathcal D_f$ contain trajectories with nonempty $\mathcal H_\zeta^f$. Only positive valid changes $\upsilon_k>0$ enter the population for $\upsilon$. The empirical distribution and quantile are
\begin{equation}
\begin{aligned}
\widehat F_f(v)&=\frac{1}{|\mathcal D_f|}\sum_{\zeta\in\mathcal D_f}
\frac{1}{|\mathcal H_\zeta^f|}\sum_{k\in\mathcal H_\zeta^f}\ind[f_k\le v],\\
\vartheta_f&=\inf\{v:\widehat F_f(v)\ge q_f\}.
\end{aligned}
\label{eq:thresholds}
\end{equation}
Here $q_f\in(0,1)$ is the quantile level. Each contributing trajectory receives equal total weight in the threshold population. Since $\log(\varepsilon_0+\chi)$ is strictly increasing, the implementation equivalently transforms the inverse CDF quantile of the gap magnitude. If the threshold population is insufficient, pre-calibration weights are retained.

Teacher responses use the native prompt and parser for each benchmark with native thinking mode disabled. Each complete response, including any visible rationale, replaces the student's original response in continuation history. Empty or unparseable responses and those yielding inadmissible or unchanged actions are excluded. Within a fixed proposal budget, candidates are checked chronologically. The first replay-valid turn with a legal teacher action different from the student's original action is selected for paired student continuations. Independent environment instances replay the task prefix and check the intended observation and action context. Both continuations use the same frozen current student, remaining budget, decoding settings, and seed schedule at each step. Replay and runtime failures invalidate the comparison rather than count as task failures.

Only tokens from the student's own trajectories enter the loss. The frozen rollout version supplies the importance ratio denominator, and asynchronous collection retains the existing staleness control. Both paired student continuations use that same student version. Their final task outcomes determine the additional weight in Equation~\ref{eq:weight}; the original token-level teacher supervision remains the signed coefficient in the loss.

\section{Experimental details}
\label{app:evidence}
\subsection{Model configurations and comparison scope}
\label{app:configurations}
The model pairings follow Section~\ref{sec:setup}. Runs use one node with $8\times$A800 GPUs. The empirical analysis, gap magnitude ranking, ablations, and timing use the Qwen3-32B teacher and Qwen3-1.7B student. The WebShop training dynamics in Section~\ref{sec:training-analysis} cover both configurations.

Table~\ref{tab:ogopd-hyperparameters} summarizes the OG-OPD weighting and calibration settings. Complete training configurations are provided with the supplementary source code.

\begin{table}[t]
\centering
\caption{\textbf{OG-OPD weighting and calibration settings.} Candidate thresholds follow Equation~\ref{eq:thresholds}. Training uses one matched pair per selected position, whereas the empirical analysis uses five matched trials per decision.}
\label{tab:ogopd-hyperparameters}
\small
\setlength{\tabcolsep}{5pt}
\renewcommand{\arraystretch}{1.08}
\begin{tabular*}{\linewidth}{@{\extracolsep{\fill}}ll}
\toprule
\textbf{Parameter} & \textbf{Value}\\
\midrule
Pre-calibration weight cap $\beta_{\max}$ & $1.2$\\
Positive-outcome weight floor $\beta_{\uparrow}$ & $1.5$\\
Gap-magnitude / positive-increase quantile & $q_\nu=0.5$ / $q_\upsilon=0.6$\\
Candidate checks / teacher proposals per trajectory & At most $2$ / $2$\\
Paired-evaluation positions per trajectory & At most $1$\\
Matched trials per selected position & $1$ pair\\
\bottomrule
\end{tabular*}
\end{table}

Figure~\ref{fig:dynamics} reports WebShop training curves separately from table results averaged over three seeds. Entropy and Teacher NLL are averaged over five steps, and the remaining rates are pooled over five batches.

The RL teachers are obtained by training Qwen3-8B separately on ALFWorld and WebShop with GiGPO \citep{feng2025gigpo}, following the teacher training approach of \citet{chen2026futurebridge}. Both use AdamW with learning rate $10^{-6}$ and weight decay $0.01$. The discount factor is $0.95$, the rollout group size is 8, and the invalid action penalty coefficient is $0.1$. Training batch sizes are 32 for ALFWorld and 16 for WebShop, with a PPO minibatch size of 64. Teacher training uses a prompt length limit of 10,240 tokens, with interaction limits of 30 steps for ALFWorld and 15 steps for WebShop. Responses are limited to 512 tokens, with thinking mode disabled. Training and validation temperatures are 1.0 and 0.4, respectively. We use the ALFWorld checkpoint at step~150 and the WebShop checkpoint at step~100, held fixed across all compared methods.

Performance and ablation comparisons evaluate complete training procedures at step~200. Ablations retain each procedure's resulting number of additionally upweighted turns and total supervision weight, rather than matching these quantities across variants. OG-OPD includes additional environment execution and final task outcome feedback, so these comparisons assess the complete algorithm. They use a fixed number of optimization steps, with actual training costs reported separately in Appendix~\ref{app:efficiency}.

\subsection{Evaluation tasks, metrics, and aggregation}
\label{app:details}
Following \citet{wang2026tcod}, ALFWorld evaluation uses 140 Seen tasks in environments encountered during training and 134 Unseen tasks with room layouts and object combinations not encountered during training. We also use TCOD's fixed Hard collection of 121 tasks drawn from the training split, on which its teacher failed in all ten sampling attempts. We report SR for each collection and Overall SR over all 395 tasks pooled together. ScienceWorld evaluates 1,308 canonical tasks. WebShop uses 100 held-out sessions, as in \citet{chen2026futurebridge}. Interaction limits are 30 steps for ALFWorld and ScienceWorld and 15 steps for WebShop. ScienceWorld success requires full completion. WebShop uses the shared TCOD success criterion of terminal reward above 0.5.

For episode $j$, let $s_{j,t}$ be its ScienceWorld environment score on a 0--100 scale and $r_{j,\mathrm{end}}$ its WebShop terminal reward on a 0--1 scale. Score is computed for each benchmark as
\begin{equation}
\mathrm{Score}_{\mathrm{SW}}=\frac{1}{n}\sum_{j=1}^{n}\operatorname{clip}\!\left(\max_t s_{j,t},0,100\right),
\quad
\mathrm{Score}_{\mathrm{WS}}=\frac{100}{n}\sum_{j=1}^{n}r_{j,\mathrm{end}}.
\label{eq:benchmark-scores}
\end{equation}
Rounds averages the actual interaction lengths of all evaluated episodes, including failures. Score is extracted within each episode as defined above, using the maximum ScienceWorld score and the terminal WebShop reward.

For episode success label $Y_j$ under the criterion for that benchmark and executed interaction length $L_j$, the remaining aggregates are
\begin{equation}
\mathrm{SR}=\frac{100}{n}\sum_{j=1}^{n}Y_j,\qquad
\mathrm{Rounds}=\frac{1}{n}\sum_{j=1}^{n}L_j,\qquad
\mathrm{Overall}_{\mathrm{ALF}}=\frac{\sum_{s\in\mathcal S}n_s\,\mathrm{SR}_s}{\sum_{s\in\mathcal S}n_s},
\label{eq:benchmark-aggregates}
\end{equation}
where $\mathcal S=\{\mathrm{Seen},\mathrm{Unseen},\mathrm{Hard}\}$ and $n_s$ is the number of evaluated tasks in collection $s$. Overall SR gives each task equal weight.

Trained-model results in the main and ablation tables use the fixed endpoint at optimizer step~200 and report means and sample standard deviations across three seeds. For any evaluation metric $M$, we first compute the metric over the benchmark's episodes for each seed, then report
\begin{equation}
\overline M_{200}=\frac{1}{3}\sum_{r=1}^{3}M_{r,200},\qquad
s_{M,200}=\sqrt{\frac{1}{3-1}\sum_{r=1}^{3}\bigl(M_{r,200}-\overline M_{200}\bigr)^2}.
\label{eq:seed-aggregation}
\end{equation}
Each trained model is evaluated at the fixed endpoint, and zero-shot references are evaluated without training. Timing is measured per run and averaged across seeds as described in Appendix~\ref{app:efficiency}.

\section{Training efficiency}
\label{app:efficiency}
\begin{table}[htbp]
\caption{\textbf{Training cost of vanilla OPD and OG-OPD.} Runs use $8\times$A800 GPUs with the Qwen3-32B teacher and Qwen3-1.7B student. OG-OPD timing includes outcome-based calibration, and Cost is normalized to vanilla OPD. Rounds reports mean evaluation interaction rounds. Boldface marks the best value per benchmark.}
\label{tab:cost}
\centering
\setlength{\tabcolsep}{3pt}
\small
\begin{tabular*}{\linewidth}{@{\extracolsep{\fill}}llccc}
\toprule
Benchmark & \textbf{Method} & Rounds $\downarrow$ & s/step $\downarrow$ & Cost $\downarrow$\\
\midrule
ScienceWorld & Vanilla OPD & 12.6 & \textbf{156.97} & \textbf{$1.00\times$}\\
 & \textbf{OG-OPD} & \textbf{11.6} & 223.58 & $1.42\times$\\
WebShop & Vanilla OPD & 6.8 & \textbf{67.21} & \textbf{$1.00\times$}\\
 & \textbf{OG-OPD} & \textbf{5.6} & 99.70 & $1.48\times$\\
\bottomrule
\end{tabular*}
\end{table}

For each run, elapsed time from training start through step~200 is divided by 200, then averaged across three seeds. The measurement includes state replay, teacher response generation, and paired student continuations, and excludes final synchronization and evaluation. Deployment uses only the trained student.

Let $T_{m,e,r}$ denote this measured elapsed time for method $m$, benchmark $e$, and training seed $r$. With $K=200$ optimization steps and $R=3$ seeds, the reported seconds per step and normalized cost are
\begin{equation}
\bar\tau_{m,e}=\frac{1}{R}\sum_{r=1}^{R}\frac{T_{m,e,r}}{K},\qquad
\mathrm{Cost}_{m,e}=\frac{\bar\tau_{m,e}}{\bar\tau_{\mathrm{OPD},e}}.
\label{eq:timing-aggregation}
\end{equation}
Cost is the ratio of the two mean times per step, rather than a mean of ratios computed for individual seeds. The corresponding absolute and relative increases are
\begin{equation}
\Delta\bar\tau_e=\bar\tau_{\mathrm{OG-OPD},e}-\bar\tau_{\mathrm{OPD},e},\qquad
\eta_e=100\left(\mathrm{Cost}_{\mathrm{OG-OPD},e}-1\right).
\label{eq:timing-increase}
\end{equation}
Using the rounded times in Table~\ref{tab:cost}, these increases are 66.61 seconds per step on ScienceWorld and 32.49 seconds on WebShop, or approximately 42\% and 48\%. They describe end-to-end training overhead under the reported hardware and execution settings, not a separate timing measurement of each calibration operation.
\end{document}